\documentclass{article}

\usepackage{PRIMEarxiv}

\usepackage[utf8]{inputenc} 
\usepackage[T1]{fontenc}    
\usepackage{hyperref}       
\usepackage{url}            
\usepackage{booktabs}       
\usepackage{amsfonts}       
\usepackage{nicefrac}       
\usepackage{microtype}      
\usepackage{lipsum}
\usepackage{graphicx}
\usepackage{amsmath}

\title{How Many Samples to Label for an Application given a Foundation Model? Chest X-ray Classification Study}

\author{
  Nikolay Nechaev \\
  AIRI \\
  Moscow\\
  \texttt{nechaev@airi.net} \\
  \And
  Evgeniia Przhezdzetskaia \\
  AIRI \\
  Moscow\\
  \texttt{przhezdzetskaia@airi.net} \\
  \And
  Viktor Gombolevskiy \\
  AIRI \\
  Moscow\\
  \texttt{gombolevskiy@airi.net} \\
  \And
  Dmitry Umerenkov \\
  AIRI  \\
  Moscow\\
  \texttt{dumerenkov@airi.net} \\
  \And
  Dmitry Dylov \\
  AIRI,  Scoltech \\
  Moscow\\
  \texttt{d.dylov@gmail.com} \\
}

\begin{document}
\maketitle

\begin{abstract}
Chest X-ray classification is vital yet resource-intensive, typically demanding extensive annotated data for accurate diagnosis. Foundation models mitigate this reliance, but how many labeled samples are required remains unclear. We systematically evaluate the use of power-law fits to predict the training size necessary for specific ROC-AUC thresholds. Testing multiple pathologies and foundation models, we find XrayCLIP and XraySigLIP achieve strong performance with significantly fewer labeled examples than a ResNet-50 baseline. Importantly, learning curve slopes from just 50 labeled cases accurately forecast final performance plateaus. Our results enable practitioners to minimize annotation costs by labeling only the essential samples for targeted performance.
\end{abstract}

\keywords{Foundation models  \and Weakly supervised \and Sample size}

\section{Introduction}

While significant literature exists on deep learning methods for chest X-ray classification\cite{meedeniya2022chest}, comparatively little attention has been paid to efficient training size estimation in this context\cite{viering2022shape}. Recent progress in foundation models has further heightened the importance of this question: not only can these models achieve higher accuracy, but their learning curves may also be more predictable with fewer labeled samples. Motivated by these considerations, we propose a systematic approach to estimate how many annotated examples a given model requires to meet a clinical ROC-AUC threshold, leveraging power-law fitting to the learning curves.

\section{Related work}

Chest X-ray is a crucial diagnostic imaging modality that provides rapid and cost-effective insights into various pulmonary and cardiac conditions \cite{raoof2012interpretation}. The classification of chest X-ray pathologies is well-studied, and training machine learning models for new conditions is relatively straightforward, although it still relies on large annotated datasets to achieve clinically acceptable accuracy \cite{ccalli2021deep}. Recently, general self-supervised learning (SSL) frameworks such as DINO \cite{oquab2023dinov2} and CLIP \cite{radford2021learning} have shown great promise for imaging tasks by learning robust feature representations from massive unlabeled datasets. These frameworks differ in their training objectives: DINO is purely image-based self-supervision, whereas CLIP leverages paired text–image data for multi-modal alignment. Building on these advances, specialized chest X-ray foundation models (\textit{e.g.}, RadDINO \cite{perez2024rad}, XraySigLIP/XrayCLIP \cite{chexagent-2024}) adapt these frameworks to chest x-ray imaging.

Though no works have directly explored learning curve estimates for chest X-ray classification tasks, many investigations in other domains (\textit{e.g.}, machine translation, image recognition, and speech recognition) rely on power-law approximations to characterize how performance improves as the training set size grows \cite{cortes1993learning}, \cite{gu2001modelling}, \cite{hestness2017deep}. Nonetheless, numerous studies reveal that learning curves can be well-behaved or ill-behaved, with phenomena such as double descent and peaking complicating straightforward sample-size extrapolation \cite{raudys1998expected}, \cite{devroye2013probabilistic}, \cite{nakkiran2020optimal}. A popular strategy is to estimate the asymptotic accuracy by measuring the early slope of a power-law fit and extrapolating the eventual plateau in performance \cite{hoiem2021learning}, \cite{frey1999modeling}, \cite{kolachina2012prediction}. Additionally, a relevant idea is to incorporate progressive sampling, dynamically refining the power-law estimate of the learning curve so as to reduce annotation overhead \cite{provost1999efficient}. 

\section{Methods}
\subsection{Dataset Construction}

A popular open dataset MIMIC-CXR \cite{johnson2019mimic} was used as the data source for this study. 
Using RadGraph annotations \cite{jain2021radgraph}, we extracted structured “organ–pathology” labels.

These labels underwent a normalization to merge synonymous anatomical terms into unified categories (\textit{e.g.}, unifying the tokens “lung” and “lobe”) and then split into two groups: normal and pathological.

\begin{figure}[ht]
  \centering
  \includegraphics[page=1,width=1.0\textwidth]{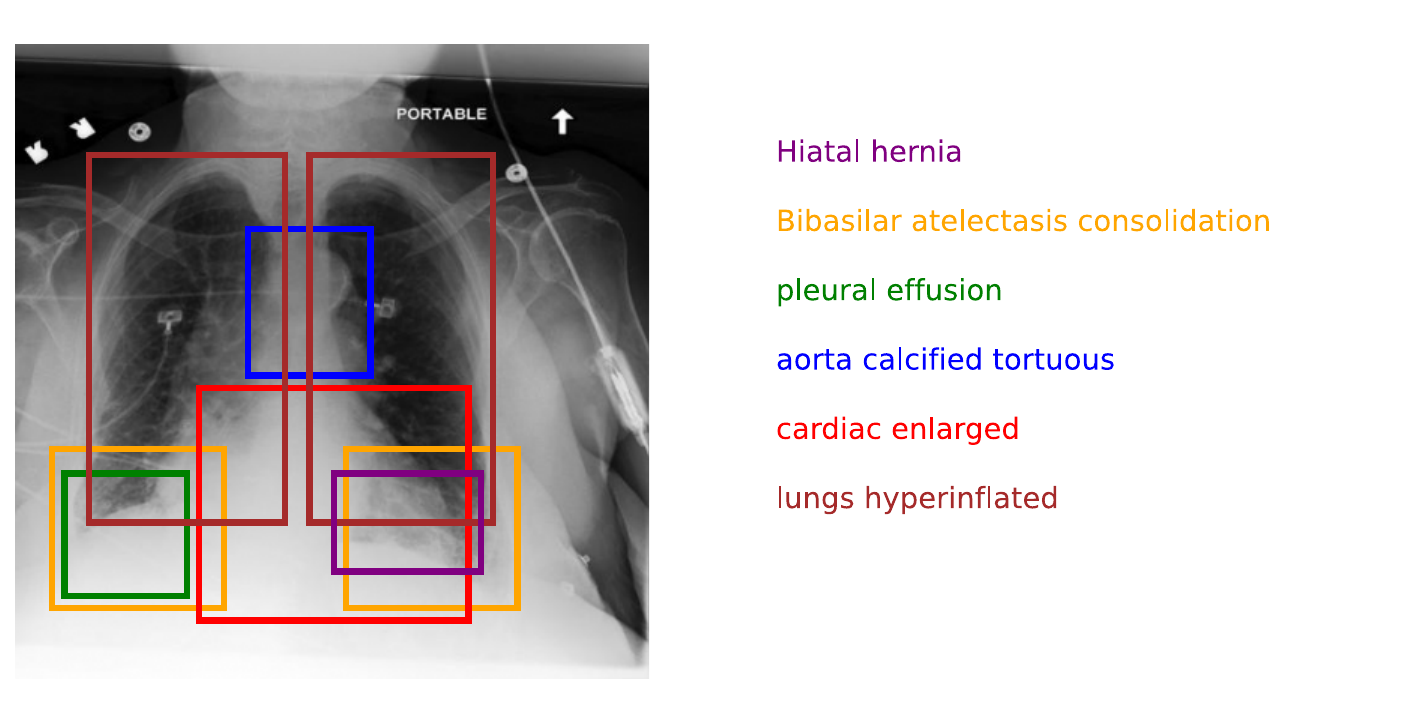}
  \caption{A chest X-ray example, with example pathologies.}
  \label{fig:data_sample}
\end{figure}

Pathological annotations were further clustered according to their common pathogenetic mechanisms to reduce redundancy. 
In total, 21 distinct pathology classes where selected.
An example of chest X-ray, corresponding RadGraph findings, and selected pathologies are shown in Figure \ref{fig:data_sample}.

For each resulting pathology, we created a binary classification dataset, where a confirmed pathology was labeled as a positive class. 
The negative class consisted of studies corresponding to a normal anatomical-physiological state of the target organ, in a 1:5 ratio.
If for some classes negative examples were insufficient, existing data were duplicated to maintain the balance.

Each pathology-specific dataset was split into training, validation, and test subsets using a deterministic method with a fixed seed value. The validation and hold-out test subsets were each assigned 10\% of the total data in a stratified manner, preserving the 1:5 class ratio. The remaining 80\% made up the full training pool.
The choice of fixed increments (ranging from 5 to 1000 samples) and the 1:5 positive-to-negative class ratio were selected to ensure comparability across a wide range of sample sizes and pathologies, and to systematically estimate minimal labeling requirements.

In each experiment below, the training sets were formed for a feature-based transfer learning regime as follows:
\begin{enumerate}
    \item 
    The number of positively annotated samples was purposely restricted with a value $N_{\text{cases}}$ taken from the set $\{5, 10, 15, ...,45, 50, 100, 250, 500, 1000\}$.
    \item From the initial training pool, $N_{cases}$ pathological studies were randomly selected (using a unique seed for each experiment).
    \item Negative cases were added in a 1:5 ratio to preserve the original balance.
\end{enumerate}

The validation and the testing sets were the same for each pathology in all experiments. The choice of fixed increments (ranging from 5 to 1000 samples) and the 1:5 positive-to-negative class ratio were selected to ensuring comparability across a wide range of sample sizes and pathologies.

\subsection{Models tested}

For feature extraction, we employed 3 different chest x-ray foundation models. 
The first, RadDINO-Maira2  \cite{bannur2024maira}, is a transformer pre-trained using the DINO-v2 framework on a heterogeneous corpus of 1.2 million medical images. The second and the third are XraySigLIP and XrayCLIP -- also transformer models, but pre-trained using the CLIP\cite{radford2021learning} and SigLIP\cite{zhai2023sigmoid} frameworks, respectively, on a million image-text pairs from CheXinstruct\cite{chexagent-2024} dataset. To verify the robustness of the results and to establish a baseline, we also used ResNet-50 convolutional neural network\cite{he2015deepresiduallearningimage} pre-trained on ImageNet as a baseline encoder. For each of the tested models we constructed the classifier head by applying a dropout layer (p = 0.1) to the encoder’s pooled features, followed by a linear projection to a single output unit.

Although labels for some pathologies are present in the MIMIC-CXR dataset, the foundation models used in our study RadDINO-Maira2, XrayCLIP and XraySigLIP were not explicitly trained using these structured pathology labels. RadDINO-Maira2 utilized only unlabeled images, while XrayCLIP and XraySigLIP were trained solely on unstructured image-text pairs without direct access to the structured pathology annotations. This ensures an unbiased evaluation, placing all pathologies on equal footing regarding the pretraining data.

Each individual experiment (pathology-model-training size) was repeated 10 times. This enabled us to assess model stability and the influence of randomness on the final metrics. 

The model with the lowest validation loss was evaluated on the hold-out test subset to determine the result of the individual experiment.

\subsection{Training Procedure}

Training was done using the \texttt{transformers} library (PyTorch backend) with the AdamW optimizer (binary cross-entropy loss, an initial learning rate $2\times 10^{-5}$), a cosine annealing learning rate scheduler without warm-up, a batch size of 64, and an early stopping after 4 consecutive epochs without improvement in the validation loss.
During training, the image encoder weights were frozen to retain their pre-trained representations, and only the linear binary classifier head was trained.

We applied train augmentations combining geometric and photometric modifications: a horizontal flip ($50\%$ probability), affine transformations with a random rotation between $-90^\circ$ and $90^\circ$ and a rotation center is center of the image size, photometric adjustments via linear brightness adaptation within $\pm 35\%$ of the original values alongside non-linear gamma correction of contrast in the same range, and spatial cropping with random square crops, covering $3$--$33\%$ of the original image size.

\subsection{Power law fitting}

To model the scaling behavior of the classifier, we fit a power law function to the area under the receiver-operating characteristic curve (ROC-AUC) in the following form:
\begin{equation}
\mathrm{ROC\_AUC}(n) = \alpha - \frac{\beta}{n^\gamma},
\end{equation}
where $n$ is the number of distinct positive examples in the training set, $\alpha$ represents the asymptotic performance, $\beta$ controls the deviation from the asymptote, and $\gamma$ governs the rate of convergence.
Multiple curve-fitting approaches (linear, exponential, and power-law) were considered. The three-parameter power-law was selected due to its consistently superior fit across the range of pathologies and models evaluated.

To estimate the parameters $\alpha$, $\beta$, and $\gamma$, we employ non-linear least squares fitting using the \texttt{curve\_fit} function from \texttt{scipy.optimize}. 
For the fitting procedure, we specify an initial guess and bounds for the parameters to ensure reasonable behavior of the model. In our case, we set the initial guess as $\alpha =  0.95,  \beta = 0.5, \gamma = 1.0 $
with the following bounds: $\alpha \in [0.8, 1.0]$ ensuring the asymptotic value is near 1, $\beta \geq 0$ to maintain non-negative deviation, $\gamma \geq 0$ for a proper convergence rate.

For the fitted power law we use the following notation ROC\_AUC$_{{N_\text{cases}}}(n)$ where $N_{\text{cases}}$ is the maximum number of examples used to fit the curve. For example, ROC\_AUC$_{20}(n)$ stands for the power law curve, fitted on the experimental data points $N_{\text{cases}} = 5, 10, 15, 20$. Finally, given the fitted curve, we draw a conclusion about the optimal number of required labeled samples $n_{o}$ by evaluating where the curve starts exceeding a certain clinically-relevant threshold (ROC\_AUC$_{{N_\text{cases}}}(n_{o})=90\%$)\footnote{The ROC-AUC threshold of 0.90 used throughout this study was selected as an illustrative benchmark for simplicity and consistency. However, our methodology is generalizable and can readily accommodate any clinically relevant performance threshold, allowing practitioners to adjust the labeling requirements according to specific diagnostic standards.}.

\begin{table}[ht]
\centering
\begin{tabular}{lcc|cc|cc|cc}
\toprule
 & \multicolumn{2}{c}{ResNet-50} & \multicolumn{2}{c}{Rad-DINO} & \multicolumn{2}{c}{XrayCLIP} & \multicolumn{2}{c}{XraySigLIP} \\
\cmidrule(lr){2-3} \cmidrule(lr){4-5} \cmidrule(lr){6-7} \cmidrule(lr){8-9}
Pathology & roc & n@90 & roc & n@90 & roc & n@90 & roc & n@90 \\
\midrule
pulmonary\_fibrosis & 0.85 & 2545 & 0.92 & 104 & 0.97 & 24 & \textbf{0.99} & \textbf{8} \\
pericardial\_effusion & 0.65 & \(>\)1M & 0.73 & 98922 & 0.77 & 5486 & \textbf{0.92} & \textbf{79} \\
aortic\_dissection & 0.72 & \(>\)1M & 0.81 & 241 & 0.71 & 4656 & \textbf{0.98} & \textbf{18} \\
hiatal\_hernia & 0.78 & \(>\)1M & 0.92 & 646 & 0.90 & 317 & \textbf{0.93} & \textbf{120} \\
lobe\_mass & 0.71 & \(>\)1M & 0.81 & 156K & 0.86 & 7605 & \textbf{0.96} & \textbf{53} \\
hemidiaphragm\_eventration & 0.78 & 1805 & 0.84 & 5315 & \textbf{0.86} & \textbf{688} & 0.84 & 8954 \\
fissure\_fluid & 0.64 & inf & 0.67 & 162K & 0.75 & 246K & \textbf{0.95} & \textbf{45} \\
spine\_deformities & 0.81 & 1159 & 0.75 & \(>\)1M & 0.87 & \(>\)1M & \textbf{0.91} & \textbf{77} \\
pulmonary\_hypertension & 0.56 & \(>\)1M & 0.72 & \(>\)1M & 0.70 & \(>\)1M & \textbf{0.86} & \textbf{1461} \\
clavicular\_fracture & 0.56 & \(>\)1M & 0.69 & \(>\)1M & \textbf{0.74} & 172K & 0.73 & \(>\)1M \\
esophagus\_dilated & 0.45 & inf & 0.57 & \(>\)1M & 0.63 & \(>\)1M & \textbf{0.82} & \textbf{161} \\
lung\_edema & 0.85 & 510 & 0.77 & 107K &  &  & \textbf{1.00} & \textbf{15} \\
diaphragms\_flattened & 0.85 & 412 & 0.85 & 13228 & \textbf{0.93} & \textbf{233} & 0.90 & 272 \\
rib\_fractures & 0.60 & \(>\)1M & 0.73 & \(>\)1M & \textbf{0.89} & 999 & 0.87 & \textbf{92} \\
lung\_aeration & 0.67 & \(>\)1M & 0.72 & 229K & \textbf{0.96} & 49 &  &  \\
hilar\_mass & 0.69 & \(>\)1M & 0.80 & 114K & \textbf{0.91} & 306 & \textbf{0.91} & \textbf{80} \\
aorta\_calcification & 0.74 & \(>\)1M & 0.75 & 267K & 0.88 & 338 & \textbf{0.89} & \textbf{97} \\
mediastinum\_shift & 0.68 & \(>\)1M & 0.82 & 7904 & 0.95 & \textbf{18} & \textbf{0.98} & 22 \\
lung\_atelectasis & 0.64 & \(>\)1M & 0.58 & \(>\)1M & \textbf{0.89} & 7071 & 0.81 & \textbf{67} \\
pleural\_air & 0.75 & \(>\)1M & 0.85 & \(>\)1M & \textbf{0.96} & \textbf{22} & 0.95 & 34 \\
cardiac\_enlarged & 0.63 & \(>\)1M & 0.71 & \(>\)1M & \textbf{0.86} & 4439 & \textbf{0.86} & \textbf{2365} \\
\bottomrule
\end{tabular}
\caption{Performance metrics with best values highlighted in bold. Best ROC-AUC and best n@90 are shown in bold.}
\label{tab:performance_metrics}
\end{table}

\section{Results}

\subsection{All pathologies data points}

The results for all 21 pathologies are presented in table \ref{tab:performance_metrics}. For the 4 models and each of the pathologies we provide the experimental ROC-AUC on all available training data for this pathology ($N_{max}$), and the expected number of cases needed to reach ROC-AUC 0.9. This number was calculated by fitting a power law to all the experimental data using less than 50 training samples and using it to calculate the number of examples needed.

\begin{figure}
\includegraphics[page=1,width=1.0\textwidth]{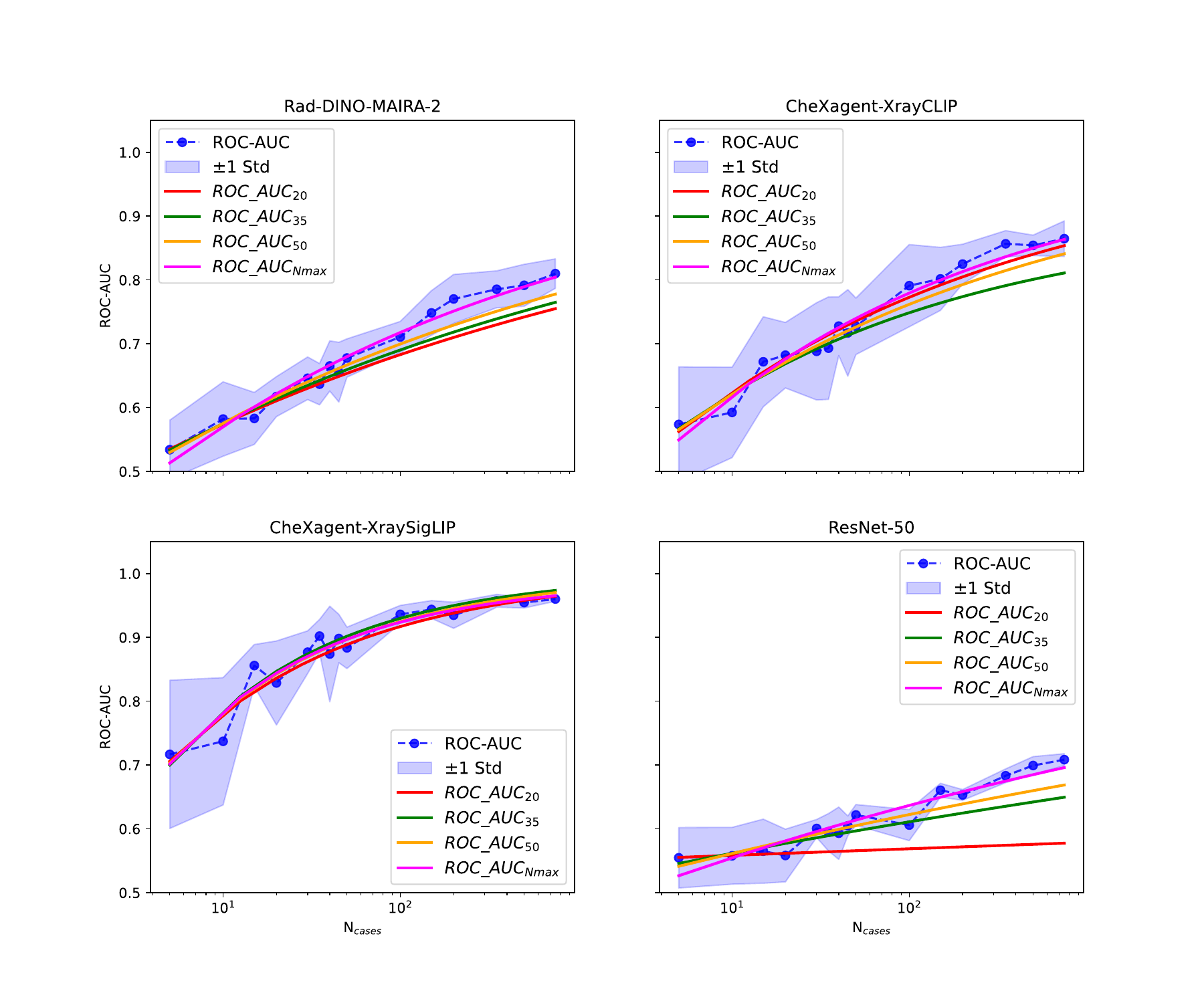}
\caption{ROC-AUC vs the number of training examples for lobe mass pathology.} \label{lobe_mass}
\end{figure}

\subsection{Number of training examples vs experimental ROC-AUC}

The experimental ROC-AUC as well as several power law estimations for an example pathology are shown in Figure \ref{lobe_mass}. For each of the four models we plot the experimental ROC-AUC vs number of different positive examples in the training set. Each point in the plot is an aggregation of 10 runs with different random seeds. We also show 4 power law fits: fitted on all experimental data and fitted on experiments with at most 20, 35 and 50 training examples.

\subsection{Comparison of Early Slope and Final Performance} 

Our first set of experiments sought to determine whether the slope of the learning curve at a relatively small training size (\textit{e.g.} n=50) could reliably predict the eventual performance plateau at large sample sizes. 

\begin{figure}
\includegraphics[page=1,width=1.0\textwidth]{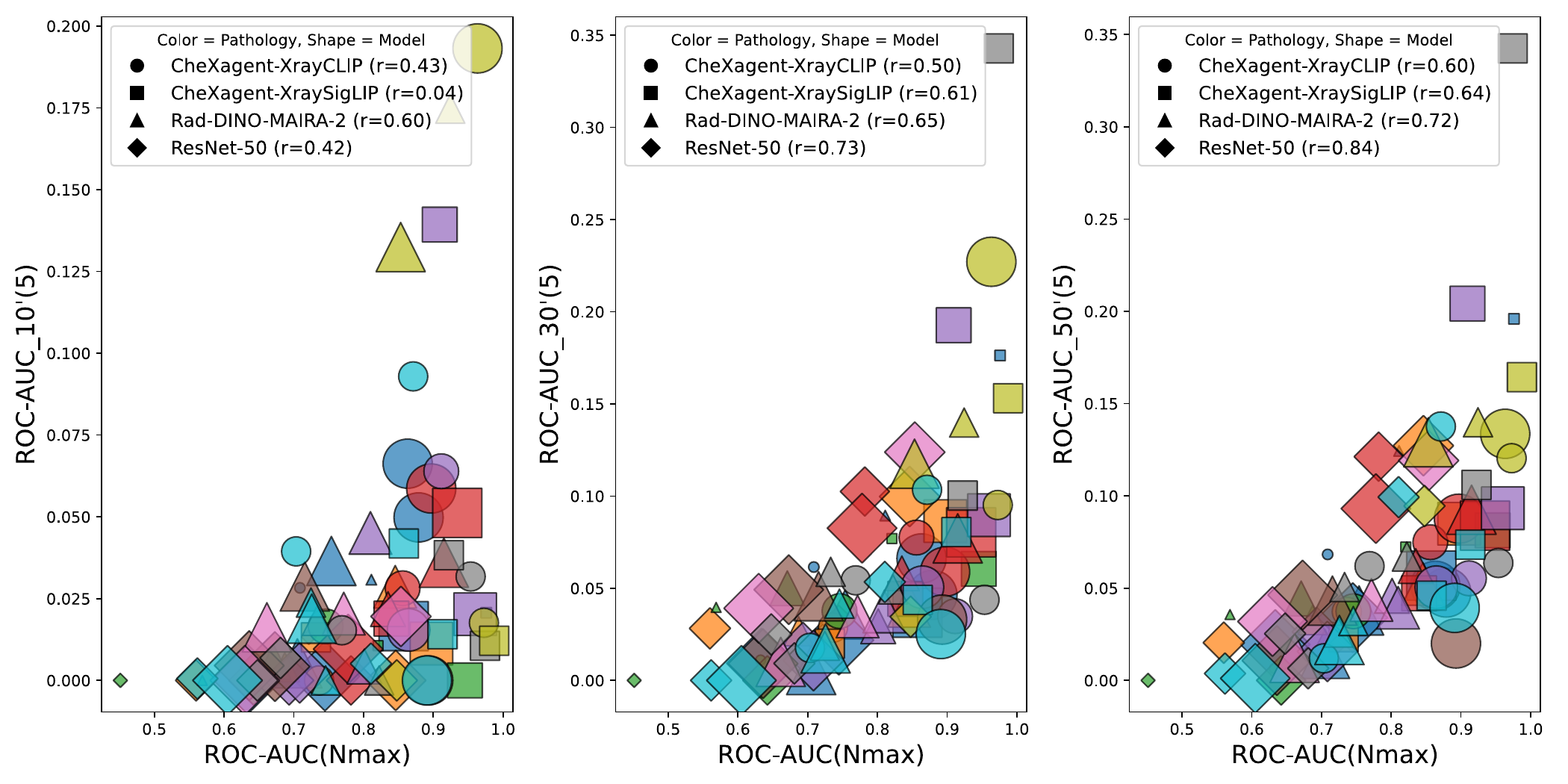}
\caption{Correlation between the derivatives of the fitted ROC-AUC at n=5 and the value of ROC-AUC(Nmax).} \label{cut_vs_err2}
\end{figure}

The slope of the learning curve is defined by ROC\_AUC$'(n) = \frac{d}{dn}\left(\alpha - \beta n^{-\gamma}\right) = \frac{\beta\gamma}{n^{\gamma+1}}$ and at n= 50 can be calculated as ROC\_AUC$'(50) = \frac{\beta\gamma}{50^{\gamma+1}}$
We fit power-law curves to the empirical ROC-AUC values, using only data from experiments with a limited number of training examples). Figure \ref{cut_vs_err2} illustrates the relationship between the steepness of the left side of the fitted slope (the derivative at n=5) and the measured ROC--AUC at Nmax. Each point on the scatter plot corresponds to one model--pathology pair, with different markers denoting the model architecture and colors indicating the specific pathology, and the marker size representing the total number of examples for this pathology. Pearson correlation \textbf{(r)} was also calculated for these values and was as expected increasing with the number of training examples. This strong correlation validates our central finding that the slope of the learning curve at early stages (small training sizes) reliably predicts the eventual performance plateau, making power-law extrapolation from small training subsets a practical tool for estimating performance at larger scales.

Figure \ref{cut_vs_err2} illustrates that the slope of the learning curve at early stages (small training sizes) strongly correlates with the eventual performance plateau, validating our central finding that power-law extrapolation from small training subsets reliably predicts performance at larger scales.

\subsection{Error in Predicted vs.\ Observed Plateau}

\begin{figure}
\centering
\includegraphics[page=1,width=0.750\textwidth]{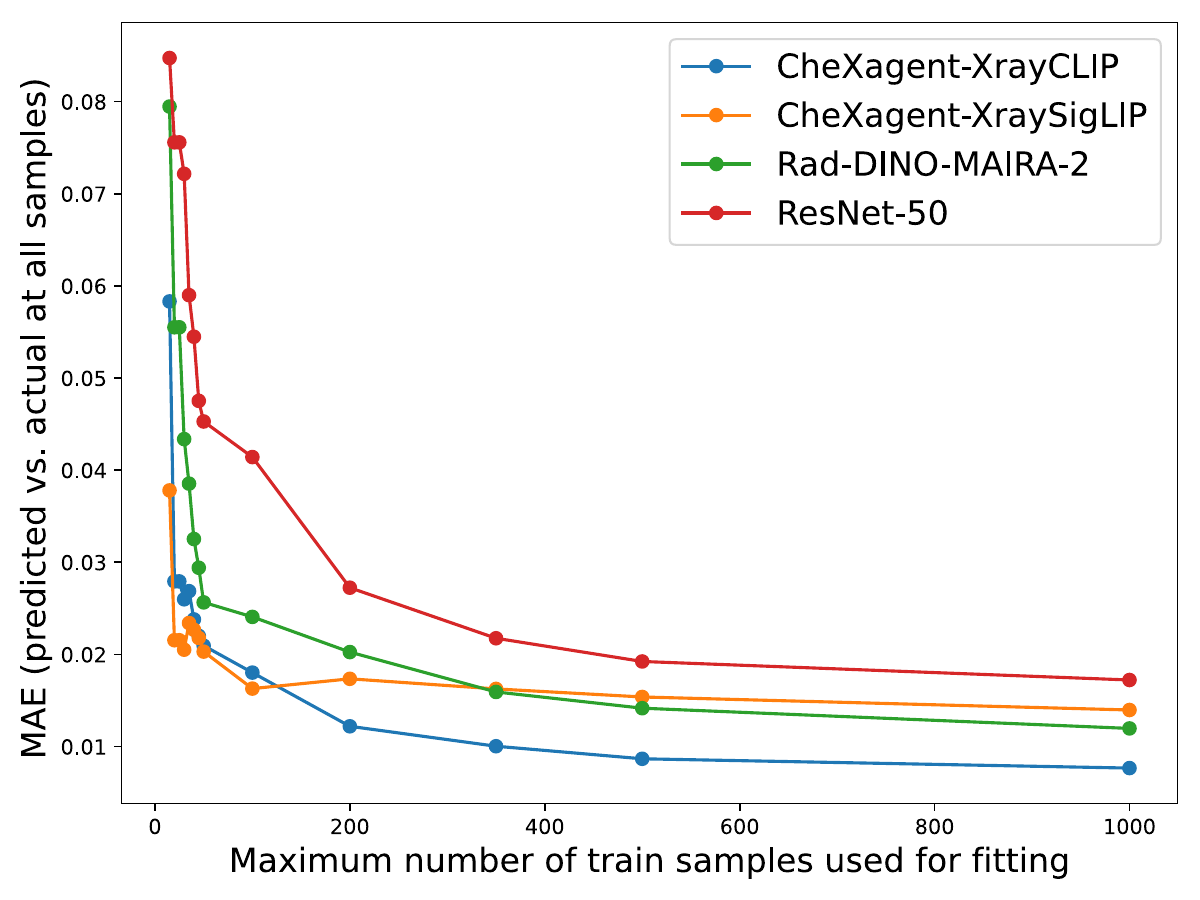}
\caption{MAE between experimental ROC-AUC and ROC-AUC predicted on limited number of training examples.} \label{cut_vs_err}
\end{figure}

Beyond measuring correlation, we also assessed absolute prediction error in estimating the final ROC-AUC. For each model-pathology pair, we used the power-law curve fitted at ${N_\text{cases}}=20$ and ${N_\text{cases}}=40$ to \emph{extrapolate} to ${{N_\text{max}}}$ equal to the maximum number of examples for this pathology. We then compared this predicted $ROC\_AUC_{{N_\text{cases}}}({{N_\text{max}}})$ with the actual measured value $ROC\_AUC({{N_\text{max}}})$.
Figure~ \ref{cut_vs_err} depicts how the mean absolute error (MAE) across pathologies and models evolves as we gradually increase the cutoff for fitting. Notably, the MAE decreases rapidly up to about 50-100 labeled cases, after which the benefit of additional data for partial fits diminishes. 

\section{Discussion}
In our experiments, we demonstrate that a straightforward process of running multiple training subsets (5 to 50 positive cases, with negative samples at a fixed ratio) allows for reliable fitting of power-law curves. By examining the initial slope and partial plateaus of these curves, we can extrapolate the performance of fine-tuned foundation models for higher training sizes. This approach is particularly useful in real-world settings where annotation is expensive, since it indicates when additional labeling provides diminishing returns. Our findings show that, in many cases, labeling on the order of 50 to 100 positive samples per pathology is sufficient to predict—and often achieve—competitive diagnostic accuracy levels.

Moreover, foundation models consistently outperform the conventional ResNet-50 baseline, underscoring not only their superior accuracy but also the improved predictability of their learning curves from limited data. Their higher initial performance effectively reduces both the total labels needed and the associated clinical costs. 

While we focused on binary classification for clarity and practicality, our framework could be extended to multi-class scenarios in future work. Similarly, though we concentrated on image-based classification, the textual modality inherent in some foundation models presents an interesting direction for future research.

We anticipate that this framework—train on subsets, fit a power law, then extrapolate to an ROC-AUC target—will inform practitioners attempting to balance annotation budgets with diagnostic performance demands when deploying chest X-ray classifiers for new pathologies.

\bibliographystyle{unsrt}  
\bibliography{references}

\end{document}